\documentclass[conference,letterpaper]{IEEEtran}

\usepackage{perlaza}
\usepackage{tikz}
\usepackage{balance}

\makeatletter
\DeclareFontFamily{U}{tipa}{}
\DeclareFontShape{U}{tipa}{m}{n}{<->tipa10}{}
\newcommand{\arc@char}{{\usefont{U}{tipa}{m}{n}\symbol{62}}}%

\newcommand{\arc}[1]{\mathpalette\arc@arc{#1}}
 
\newcommand{\arc@arc}[2]{%
  \sbox0{$\m@th#1#2$}%
  \vbox{
    \hbox{\resizebox{\wd0}{\height}{\arc@char}}
    \nointerlineskip
    \box0
  }%
}
\makeatother

\makeatletter
\renewcommand{\baselinestretch}{1}
\IEEEoverridecommandlockouts
\begin{document}
\title{Empirical Risk Minimization with Relative Entropy Regularization: Optimality and Sensitivity Analysis} 

\author{Samir M. Perlaza,  Gaetan~Bisson,  I\~{n}aki~Esnaola,  Alain~Jean-Marie, and Stefano~Rini
		\thanks{Samir M. Perlaza, and Alain Jean-Marie are with INRIA, 2004  Route des Lucioles, 06902 Sophia Antipolis, France. ($\lbrace$samir.perlaza, alain.jean-marie$\rbrace$@inria.fr)}
		\thanks{Gaetan Bisson and Samir M. Perlaza are with the Laboratoire de Math\'{e}matiques GAATI, Universit\'{e} de la Polyn\'{e}sie Fran\c{c}aise,  BP 6570, 98702 Faaa, French Polynesia. (bisson@gaati.org)}
		\thanks{ I\~{n}aki Esnaola is with the Department of Automatic Control and Systems Engineering, University of Sheffield,  Sheffield, United Kingdom. (esnaola@sheffield.ac.uk)}
		\thanks{Stefano Rini is with the Department of Electrical and Computer Engineering, National Chao Tung University,  Hsinchu, Taiwan. }
		\thanks{Samir M. Perlaza and I\~{n}aki Esnaola are also with the Department of Electrical and Computer Engineering, Princeton University, Princeton, 08544 NJ, USA.}
		\thanks{This work was supported in part by the INRIA Exploratory Action ``Information and Decision Making (IDEM)''; and in part by the Agence Nationale de la Recherche under grant number ANR-20-CE40-0013.}
	}

\maketitle
\begin{abstract}
The optimality and sensitivity of the empirical risk minimization problem with relative entropy regularization (ERM-RER) are investigated for the case in which the reference is a $\sigma$-finite measure instead of a probability measure. This generalization allows for a larger degree of flexibility in the incorporation of prior knowledge over the set of models. In this setting, the interplay of the regularization parameter, the reference measure, the risk function, and the empirical risk induced by the solution of the ERM-RER problem is characterized.
This characterization yields necessary and sufficient conditions for the existence of  regularization parameters that achieve  arbitrarily small empirical risk with arbitrarily high probability.
Additionally, the sensitivity of the expected empirical risk to deviations from the solution of the ERM-RER problem is studied. Dataset-dependent and dataset-independent upper bounds on the absolute value of the sensitivity are presented. In a special case, it is shown that the expectation (with respect to the datasets) of the absolute value of the sensitivity is upper bounded, up to a constant factor, by the square root of the lautum information between the models and the datasets.
\end{abstract}
\section{Introduction}

The problem of empirical risk minimization (ERM) \cite{vapnik1992principles}, which is strongly related to $M$-estimation \cite{Stefanski-TAS-2002}, minimum contrast estimation~\cite{bm-rcmce-93}, and sample average approximation \cite{Anton-SIAM-2002},  appears in numerous central problems in machine learning \cite{shalev2014understanding}.  
Among the most popular methods for solving the ERM problem are those based on the gradient. The stochastic gradient descent algorithm \cite{robbins1951stochastic} and its variants fall within this class of methods. See, for instance,  the literature reviews in \cite{vapnik1999overview, bottou2018optimization} and \cite{xin2020decentralized}.
Other methods for solving the ERM problem are based on constructing probability measures over the  mesurable space formed by the set of models. 
In these methods, the model is sampled from a particular probability distribution and thus, the figure of merit is the expectation of the empirical risk with respect to such probability distribution.
Methods such as Bayesian methods \cite{robert2007bayesian} and PAC-Bayesian methods \cite{haddouche2020pacbayes, guedj2019free} are typical examples. 
When a prior on the models is available, a typical choice is to regularize the original ERM problem by the relative entropy with respect to the prior. This problem is know as the ERM with relative entropy regularization (ERM-RER) and has been widely studied in the context of statistical physics \cite{zdeborova2016statistical,catoni2007pac} and information theory \cite{russo2019much, zhang2006information, asadi2018chaining, asadi2020chaining,Aminiam2021Exact}. Often, the prior is in the form of a probability distribution. Nonetheless, it has been shown that the ERM-RER problem can be generalized to the case in which priors are in the form of $\sigma$-finite measures \cite{InriaRR9454}. In particular, when the prior is, for instance, the Lebesgue measure or a counting measure, the ERM-REM problem boils down to the well known problems of ERM with differential and discrete entropy regularization, respectively. See, for instance \cite{JaynesMaxEnt1, JaynesMaxEnt2, jaakkola1999maximum}.
Interestingly, even in the case in which priors are $\sigma$-finite measures, the ERM-RER problem is shown to have a unique solution \cite{InriaRR9454}.   


The contribution of this paper is twofold. First, it introduces a notion of  optimality in probability for the ERM-RER problem, which is reminiscent to a probably approximately correct (PAC) guarantee \cite{valiant1984theory}. More specifically, a solution to the ERM-RER problem is said to be  $(\delta, \epsilon)$-optimal if the set of models that induce empirical risks smaller than $\delta$ exhibits a probability higher than  $1- \epsilon$. Necessary and sufficient conditions on the parameters of the ERM-RER problem for observing a $(\delta, \epsilon)$-optimal solution are presented.
Second, the sensitivity of the ERM-RER problem is studied for a given dataset. The sensitivity is defined as the difference between two quantities: $(a)$ the expectation of the empirical risk for a given dataset with respect to a given probability measure $P$ on the set of models; and $(b)$ The expectation of the empirical risk for such dataset with respect to the measure that is the solution to  the ERM-RER problem.  In particular, an upper bound on the absolute value of the sensitivity is presented. Such upper-bound is expressed in terms of the square root of the relative entropy of $P$ with respect to the probability measure solution to the ERM-RER problem. As a byproduct, it is shown that the expectation of the absolute value of the sensitivity with respect to the probability distribution of the datasets is bounded. In a special case, it is shown that such bound is, up to a constant term,  the square root of the lautum information \cite{palomar2008lautum} between the models and the datasets.
This result is analogous to the results in \cite{Alab-NIPS2015, pmlr-v51-russo16, xu2017information}, in which, under certain conditions, the generalization gap is upper bounded by  a term that is proportional to the square root of the mutual information between the models and the datasets. 

The paper is organized as follows. Section~\ref{SecProblemFormulation} introduces the ERM and the ERM-RER problem. The solution to the ERM-RER, introduced in \cite{InriaRR9454}, is discussed. Section~\ref{SecOptimality} introduces the notions of $\left( \delta, \epsilon \right)$-optimality, coherent measures, and consistent measures. Necessary and sufficient conditions for guaranteeing that the solution to the ERM-RER problem is $\left( \delta, \epsilon \right)$-optimal are presented. Section~\ref{SecSensitivity} introduces the notion of sensitivity and presents upper-bounds on the absolute value of the sensitivity of the ERM-RER problem. These upper-bounds can be divided into two classes, dataset-dependent bounds and dataset-independent bounds.
The former hold for a given dataset, whereas the latter hold for the expectation of the absolute value of the sensitivity with respect to the probability distribution of the data. Section~\ref{SecDiscussions} concludes the paper.

\section{Problem Formulation}\label{SecProblemFormulation}

\subsection{Empirical Risk Minimization}
Consider three sets $\set{M}$, $\set{X}$ and $\set{Y}$, with $\set{M} \subseteq \reals^{d}$ and $d \in \ints$. Let the function $f: \set{M} \times \mathcal{X} \rightarrow \mathcal{Y}$ be such that, for some $\vect{\theta}^{\star} \in \set{M}$, there exist two random variables $X$ and $Y$ that satisfy, 
\begin{equation}\label{EqTheModel}
    Y = f(\vect{\theta}^{\star}, X).
\end{equation}
The random variables $X$ and $Y$ jointly form the probability space:
\begin{equation}
\label{EqPxy}
\left( \mathcal{X}\times\mathcal{Y}, \mathscr{F}\left( \mathcal{X}\times\mathcal{Y} \right), P_{X Y}\right),
\end{equation}
where $\mathscr{F}\left( \mathcal{X}\times\mathcal{Y} \right)$ is a  $\sigma$-algebra on the set $\mathcal{X}\times\mathcal{Y}$, which is assumed to be fixed in this analysis.
The elements of the sets  $\set{M}$, $\mathcal{X}$ and $\mathcal{Y}$ are often referred to as \emph{models}, \emph{patterns} and \emph{labels}, respectively.  
A pair $(x,y) \in \mathcal{X} \times \mathcal{Y}$ is referred to as a \emph{labeled pattern} or \emph{data point} under the following condition.  
\begin{definition}[Data Point]\label{DefDataPoint}
The pair $(x, y)$ is said to be a data point if $(x, y) \in \supp P_{X Y}$.
\end{definition}
Several data points form a dataset.
\begin{definition}[Dataset]\label{DefDataSet}
Given $n$ data points, with $n \in \ints$,  denoted by $\left(x_1, y_1 \right)$, $\left( x_2, y_2\right)$, $\ldots$, $\left( x_n, y_n \right)$, a dataset is represented by the tuple  $\left(\left(x_1, y_1 \right), \left(x_2, y_2 \right), \ldots, \left(x_n, y_n \right)\right) \in \left( \mathcal{X} \times \mathcal{Y} \right)^n$.
\end{definition}  
The model $\vect{\theta}^{\star}$ in~\eqref{EqTheModel}, which is often referred to as the \emph{ground truth model}, is unknown. Given a dataset, the objective is to obtain a model $\vect{\theta} \in \set{M}$, such that,  for all patterns $x \in \mathcal{X}$, the assigned label $f(\vect{\theta}, x)$ 
 minimizes a notion of \emph{loss} or \emph{risk}.  
Let the function 
\begin{equation}\label{EqEll}
\ell: \set{Y} \times \set{Y} \rightarrow [0, +\infty)
\end{equation} 
be such that  given a data point $(x, y) \in \set{X} \times \set{Y}$, the loss or risk induced by choosing the model $\vect{\theta} \in \set{M}$
is $\ell\left( f(\vect{\theta}, x), y \right)$.  %
Often, the function $\ell$  is referred to as the \emph{loss function} or \emph{risk function}. 
In the following, it is assumed that the function $\ell$ satisfies that,  for all $y \in \set{Y}$, the loss $\ell\left( y , y\right) = 0$, which implies that correct labelling induces zero cost. 
Note that there might exist several models $\vect{\theta} \in \set{M}
\setminus\lbrace \vect{\theta}^{\star}\rbrace$ such that $\ell\left( f(\vect{\theta}, x), y\right) = 0$, which reveals the need of a large number of labeled patterns for model selection.

The \emph{empirical risk} induced by the model $\vect{\theta}$, with respect to a dataset 
\begin{subequations}\label{EqOriginalOP}
\begin{equation}\label{EqTheDataSet}
\vect{z} = \big(\left(x_1, y_1 \right), \left(x_2, y_2 \right), \ldots, \left(x_n, y_n \right)\big)  \in \left( \set{X} \times \set{Y} \right)^n,
\end{equation}  
with $n\in \ints$, is determined by the  function $\mathsf{L}_{\vect{z}}: \set{M} \rightarrow [0, +\infty)$, which satisfies  
\begin{IEEEeqnarray}{rCl}
\label{EqLxy}
\mathsf{L}_{\vect{z}} \left(\vect{\theta} \right)  & = & 
\frac{1}{n}\sum_{i=1}^{n}  \ell\left( f(\vect{\theta}, x_i), y_i\right).
\end{IEEEeqnarray}

Using this notation, the ERM problem consists of the following optimization problem
\begin{equation}\label{EqOriginalOPobjective}
\min_{\vect{\theta} \in \set{M}} \mathsf{L}_{\vect{z}} \left(\vect{\theta} \right),
\end{equation}
whose solutions form the set denoted by
\begin{equation}\label{EqHatTheta}
\set{T}\left( \vect{z} \right) \triangleq \arg\min_{\vect{\theta} \in \set{M}}    \mathsf{L}_{\vect{z}} \left(\vect{\theta} \right).
\end{equation}
\end{subequations}
The ground truth model $\vect{\theta^{\star}}$ in~\eqref{EqTheModel} is one of the solutions to the ERM problem in~\eqref{EqOriginalOP}. That is, the model $\vect{\theta^{\star}}$ in~\eqref{EqTheModel} satisfies that $\vect{\theta^{\star}} \in \set{T}\left( \vect{z} \right)$ and $\mathsf{L}_{\vect{z}}\left( \vect{\theta^{\star}} \right) = 0$. Hence, the ERM problem in~\eqref{EqOriginalOP} is well posed. 
 
\subsection{Notation and Assumptions} 

The \emph{generalized relative entropy} is defined below as the extension to $\sigma$-finite measures of the relative entropy usually defined for probability measures.

%
\begin{definition}[Generalized Relative Entropy]\label{DefRelEntropy}
Given two $\sigma$-finite measures $P$ and~$Q$ on the same measurable space, such that $Q$ is absolutely continuous with respect to $P$,  the relative entropy of $Q$ with respect to $P$ is
\begin{equation}
\KL{Q}{P} = \int \frac{\mathrm{d}Q}{\mathrm{d}P}(x)  \log\left( \frac{\mathrm{d}Q}{\mathrm{d}P}(x)\right)  \mathrm{d}P(x),
\end{equation}
where the function $\frac{\mathrm{d}Q}{\mathrm{d}P}$ is the Radon-Nikodym derivative of $Q$ with respect to $P$.
\end{definition}
In the following, given a measurable space $\left( \Omega , \mathscr{F} \right)$, the notation $\triangle\left( \Omega , \mathscr{F} \right)$ is used to represent the set of $\sigma$-finite measures  that can be defined over such a measurable space. Given a measure $Q \in \triangle\left( \Omega , \mathscr{F} \right)$, the subset $\triangle_{Q}\left( \Omega , \mathscr{F} \right)$ contains all measures that are absolutely continuous with respect to the measure $Q$. Given a set $\set{A} \subset \reals^d$, with $d \in \ints$, the Borel $\sigma$-field over $\set{A}$ is denoted by $\BorSigma{\set{A}}$.

A fundamental assumption in this work is that the function $\bar{\ell}: \set{M} \times \set{X} \times \set{Y} \rightarrow [0,+\infty)$, such that for all  $\left(\vect{\theta}, x, y\right) \in \set{M}\times\set{X}\times\set{Y}$, 
\begin{equation}
\label{EqBarEll}
\bar{\ell}\left(\vect{\theta}, x, y\right) = \ell\left( f(\vect{\theta}, x), y\right),
\end{equation} 
where the functions $f$ and $\ell$ are those in~\eqref{EqTheModel} and~\eqref{EqEll}, is Borel measurable with respect to the measure space $\big( \set{M} \times \mathcal{X}\times\mathcal{Y}$, $\BorSigma{\set{M}} \times \mathscr{F}\left(\mathcal{X}\times\mathcal{Y} \right)\big)$.
%

\subsection{Generalized Relative Entropy Regularization}

Under the assumptions above, when models are chosen by \emph{sampling} from a probability measure over the measurable space $\Bormeaspace{\set{M}}$, one of the performance metrics is 
the \emph{expected empirical risk}, which is introduced hereunder.
\begin{definition}[Expected Empirical Risk]\label{DefEmpiricalRisk}
Given a dataset $\vect{z} \in \left( \set{X} \times \set{Y}\right)^n$, let the function $\mathsf{R}_{\vect{z}}: \triangle\Bormeaspace{\set{M}} \rightarrow \left[0, +\infty \right)$ be  such that for all $\sigma$-finite measures $P \in \triangle\Bormeaspace{\set{M}}$, it holds that
\begin{equation}
\label{EqRxy}
\mathsf{R}_{\vect{z}}\left( P  \right) = \int \mathsf{L}_{ \vect{z} } \left(\vect{\theta} \right)  \mathrm{d} P(\vect{\theta}),
\end{equation}
where the function $\mathsf{L}_{\vect{z}}$ is in~\eqref{EqLxy}. 
Then, when $P$ is a probability measure, the expected empirical risk induced by $P$ is $\mathsf{R}_{\vect{z}}\left( P  \right)$. 
\end{definition}

The ERM-RER problem is parametrized by a $\sigma$-finite measure in $\triangle\Bormeaspace{\set{M}}$ and a positive real, which are referred to as the \emph{reference measure} and the \emph{regularization factor}, respectively.
Let $Q \in \triangle\Bormeaspace{\set{M}}$ be a $\sigma$-finite measure  and let $\lambda$ be a positive real. The ERM-RER problem, with parameters $Q$ and $\lambda$, consists of the following optimization problem:
\begin{subequations}\label{EqERMRER}
\begin{IEEEeqnarray}{rcl}
    \min_{P \in \triangle_{Q}\Bormeaspace{\set{M}}} & &  \mathsf{R}_{\vect{z}} \left( P \right)  + \lambda D\left( P \|Q\right),\\
    \mathrm{s.t.} & & \int \d P(\vect{\theta}) = 1.
\end{IEEEeqnarray}
\end{subequations}
where the dataset $\vect{z}$ is in~\eqref{EqTheDataSet}; and the function $\mathsf{R}_{\vect{z}}$ is defined in~\eqref{EqRxy}.

The solution to the ERM-RER problem in~\eqref{EqERMRER} is presented by the following lemma.
\begin{lemma}[Theorem $2.1$ in \cite{InriaRR9454}]\label{TheoremOptimalModel}
Given a $\sigma$-finite measure $Q \in  \triangle\Bormeaspace{\set{M}}$ and  a dataset $\vect{z} \in \left( \set{X} \times \set{Y} \right)^n$, 
let the function $K_{Q,\vect{z}}: \reals \rightarrow \reals \cup \lbrace +\infty\rbrace$ be such that for all $t \in \reals$,
\begin{IEEEeqnarray}{rcl}
\label{EqK}
K_{Q,\vect{z}}\left(t \right) & = &  \log\left( \int \exp\left( t \; \mathsf{L}_{\vect{z}}\left(\vect{\theta}\right)  \right) \mathrm{d}Q(\vect{\theta}) \right),
\end{IEEEeqnarray} 
where the function $\mathsf{L}_{\vect{z}}$ is defined in~\eqref{EqLxy}.
Let also the set $\set{K}_{Q,\vect{z}} \subset \reals$ be 
\begin{IEEEeqnarray}{rcl}
\label{EqSetKxy}
\set{K}_{Q,\vect{z}} & \triangleq &\left\lbrace s \in (0, +\infty): \; K_{Q,\vect{z}}\left(-\frac{1}{s} \right)  < +\infty \right\rbrace.
\end{IEEEeqnarray}
Then, for all $\lambda \in \set{K}_{Q,\vect{z}}$, the solution to the ERM-RER problem in~\eqref{EqERMRER}, denoted by 
$P^{\left(Q, \lambda\right)}_{\vect{\Theta}| \vect{Z} = \vect{z}} \in \triangle_{Q}\Bormeaspace{\set{M}}$,  is a unique probability measure whose Radon-Nikodym derivative with respect to $Q$ satisfies for all $\vect{\theta} \in \supp Q$,
\begin{IEEEeqnarray}{rCl}\label{EqGenpdf}
\frac{\mathrm{d}P^{\left(Q, \lambda\right)}_{\vect{\Theta}| \vect{Z} = \vect{z}}}{\mathrm{d}Q} \left( \vect{\theta} \right) 
  & =& \exp\left( - K_{Q,\vect{z}}\left(- \frac{1}{\lambda} \right) - \frac{1}{\lambda} \mathsf{L}_{\vect{z}}\left( \vect{\theta}\right)\right).
\end{IEEEeqnarray}
\end{lemma}

 \section{$(\delta, \epsilon)$-Optimality}\label{SecOptimality}

This section introduces the notion of $(\delta, \epsilon)$-optimality. In particular, the focus is on the conditions on the empirical risk function $\mathsf{L}_{\vect{z}}$ in~\eqref{EqLxy} and the parameters $Q$ and $\lambda$ of the ERM-RER problem in~\eqref{EqERMRER} for observing expected empirical risks that are arbitrarily small with arbitrarily high probability.
\begin{definition} \label{DefDeltaEpsilonOptimal}
Given a pair $(\delta,\epsilon) \in [0,+\infty) \times (0,1)$, the probability measure $P^{\left(Q, \lambda\right)}_{\vect{\Theta}| \vect{Z}=\vect{z}}$ in~\eqref{EqGenpdf}, which is the solution to the ERM-RER problem in~\eqref{EqERMRER}, is said to be $(\delta, \epsilon)$-optimal, if the set
\begin{IEEEeqnarray}{rCl}
\label{EqSetL}
\set{L}_{\vect{z}} \left( \delta \right)  & \triangleq&  \left\lbrace  \vect{\theta}  \in \set{M}:   \mathsf{L}_{\vect{z}}\left( \vect{\theta} \right)  \leqslant \delta  \right\rbrace,
\end{IEEEeqnarray}
with  the function $\mathsf{L}_{\vect{z}}$ in~\eqref{EqLxy}, satisfies
\begin{equation}
\label{EqOptimalConcentration}
P^{\left(Q, \lambda \right)}_{\vect{\Theta}| \vect{Z} = \vect{z}} \left(\set{L}_{\vect{z} } \left( \delta \right) \right) > 1 - \epsilon.
\end{equation}
\end{definition}
For all $\delta>0$, it holds that $\set{T}\left( \vect{z} \right) \subset  \set{L}_{\vect{z}} \left( \delta \right)$, with the sets $\set{T}\left( \vect{z} \right)$ and $ \set{L}_{\vect{z}}$  in~\eqref{EqHatTheta} and~\eqref{EqSetL}, respectively. 
Hence, from Definition~\ref{DefDeltaEpsilonOptimal}, it follows that the probability measure $P^{\left( Q, \lambda \right)}_{\vect{\Theta}| \vect{Z} = \vect{z}}$ assigns  probability $(1 - \epsilon)$ to a set that contains the models that induce an empirical risk smaller than or equal to $\delta$.  
In view of this, it is interesting to identify the conditions on the parameters $Q$ and $\lambda$ for which the solutions to the ERM-RER problem in~\eqref{EqERMRER} are $\left(\delta , \epsilon \right)$-optimal. In the following, such conditions are stated using the following definitions.
%
%
%
\begin{definition}\label{DefCoherentMess}
The $\sigma$-finite measure $Q \in \triangle\Bormeaspace{\set{M}}$ in~\eqref{EqERMRER} is said to be coherent if, for all $\delta > 0$, it holds that
\begin{equation}
Q \left( \set{L}_{\vect{z}} \left( \delta \right)  \right) > 0,
\end{equation}
where the set $\set{L}_{\vect{z}} \left( \delta \right)$ is defined in~\eqref{EqSetL}.
\end{definition}

In the case in which the $\sigma$-finite measure $Q$ in~\eqref{EqERMRER} is coherent, the probability measure $P^{\left(Q, \lambda\right)}_{\vect{\Theta}| \vect{Z} = \vect{z}}$ in~\eqref{EqGenpdf} satisfies  for all $\delta > 0$,
\begin{equation}
P^{\left(Q, \lambda\right)}_{\vect{\Theta}| \vect{Z} = \vect{z}} \left( \set{L}_{\vect{z}} \left( \delta \right)  \right) > 0.
\end{equation}  
This is a consequence of the fact that the measures $P^{\left(Q, \lambda\right)}_{\vect{\Theta}| \vect{Z} = \vect{z}}$ and $Q$ are mutually absolutely continuous \cite[Lemma $2.6$]{InriaRR9454}. 
On the other hand, when the measure $Q$ is noncoherent, it follows that $Q\left( \set{T}\left( \vect{z} \right) \right) = 0$, which implies $P^{\left(Q, \lambda\right)}_{\vect{\Theta}| \vect{Z} = \vect{z}}\left( \set{T}\left( \vect{z} \right) \right) = 0$.

\begin{definition}\label{DefConsistentMess}
The $\sigma$-finite measure $Q \in \triangle\Bormeaspace{\set{M}}$ in~\eqref{EqERMRER} is said to be consistent, if the set 
\begin{IEEEeqnarray}{rcl}
\label{EqSetLStar}
\set{L}^{\star}_{\vect{z}}  &  \triangleq &  \left\lbrace  \vect{\theta}  \in \set{M}:   \mathsf{L}_{\vect{z}}\left( \vect{\theta} \right)  =  \delta^{\star}\right\rbrace
\end{IEEEeqnarray} 
satisfies $Q\left( \set{L}^{\star}_{\vect{z}}  \right) > 0$, where  the function $\mathsf{L}_{\vect{z}}$ is in~\eqref{EqLxy}; and
\begin{equation}\label{EqDeltaStar}
\delta^{\star} \triangleq \inf\left\lbrace \delta \in [0, +\infty) : Q\left( \set{L}_{\vect{z}} \left( \delta \right) \right) > 0 \right\rbrace.
\end{equation}
\end{definition}
Note that when $Q$ is coherent, $\delta^{\star} = 0$, and thus, $ \set{L}^{\star}_{\vect{z}} = \set{T}\left( \vect{z} \right)$, with $\set{T}\left( \vect{z} \right)$ in \eqref{EqHatTheta}. Moreover, if $Q$ is coherent and consistent, then $Q\left( \set{T}\left( \vect{z} \right) \right) > 0$.
Using the elements above, the main result of this section is presented by the following theorem.

\begin{theorem}\label{TheoGibbsDeltaEps} 
If the $\sigma$-finite measure $Q \in \triangle\Bormeaspace{\set{M}}$ in~\eqref{EqERMRER} is consistent, then for all $(\delta, \epsilon) \in (\delta^{\star},+\infty) \times (0,1)$, with $\delta^{\star}$ in~\eqref{EqDeltaStar}, there always exists a $\lambda \in \set{K}_{Q, \vect{z}}$, with $\set{K}_{Q, \vect{z}}$ in~\eqref{EqSetKxy}, such that the measure $P^{\left(Q, \lambda\right)}_{\vect{\Theta}| \vect{Z}=\vect{z}}$ in~\eqref{EqGenpdf} is $(\delta, \epsilon)$-optimal.
\end{theorem}
\begin{IEEEproof}
The proof is presented in \cite[Theorem $3.1$]{InriaRR9454}.
\end{IEEEproof}        

A stronger optimality claim can be obtained when the reference measure is consistent and coherent, as shown by the following corollary of Theorem~\ref{TheoGibbsDeltaEps}.

\begin{corollary}\label{TheoGibbsCoherent}
If the $\sigma$-finite measure $Q \in \triangle\Bormeaspace{\set{M}}$ in~\eqref{EqERMRER} is coherent and consistent, then,  for all $(\delta, \epsilon) \in (0,+\infty) \times (0,1)$, there exists a $\lambda \in \set{K}_{Q, \vect{z}}$, with $\set{K}_{Q, \vect{z}}$ in~\eqref{EqSetKxy}, such that the measure $P^{\left(Q, \lambda\right)}_{\vect{\Theta}| \vect{Z} = \vect{z}}$ in~\eqref{EqGenpdf} is $(\delta, \epsilon)$-optimal.
\end{corollary}
 
\section{Sensitivity}\label{SecSensitivity}
A performance metric to evaluate the deviations of the expected empirical risk $\mathsf{R}_{\vect{z}}$ (Definition~\ref{DefEmpiricalRisk}) from the probability measure $P^{\left(Q, \lambda\right)}_{\vect{\Theta}| \vect{Z} = \vect{z}}$ in~\eqref{EqGenpdf} towards an alternative probability measure $P$ is the sensitivity, which was introduced in \cite{InriaRR9454}. 
Deviations from the probability measure $P^{\left(Q, \lambda\right)}_{\vect{\Theta}| \vect{Z} = \vect{z}}$ towards an alternative probability measure $P$ over the measurable space $\Bormeaspace{\set{M}}$ might arise due to several reasons. For instance, if new datasets become available, a new ERM-RER problem can be formulated using a larger dataset obtained by aggregating the old and the new datasets \cite{InriaRR9474}. 
Similarly, the parameters $Q$ (the reference measure) and $\lambda$ (the regularization factor) in~\eqref{EqERMRER} might be changed based on side-information leading to new ERM-RER problems and thus, to new probability measures.
Other techniques different from ERM-RER might also be used to obtain a probability measure  over the measurable space $\Bormeaspace{\set{M}}$, e.g., Bayesian methods. 
Within this context, the sensitivity is a performance metric defined as follows.
\begin{definition}[Sensitivity]
Given a $\sigma$-finite measure $Q \in \triangle\Bormeaspace{\set{M}}$ and a positive real $\lambda > 0$, let $\mathsf{S}_{Q, \lambda}: \left( \set{X} \times \set{Y} \right)^n \times \triangle_{Q}\Bormeaspace{\set{M}}\rightarrow \left( - \infty, +\infty \right]$ be a function such that for all datasets $\vect{z} \in  \left( \set{X} \times \set{Y} \right)^n$ and for probability measures $P \in \triangle_{Q}\Bormeaspace{\set{M}}$, it holds that
\begin{IEEEeqnarray}{l}
\label{EqDefSensitivity}
\mathsf{S}_{Q, \lambda}\left( \vect{z}, P \right)  =
\left\lbrace
\begin{array}{cl}
\mathsf{R}_{\vect{z}}\left( P \right)  - \mathsf{R}_{\vect{z}}\left( P^{\left(Q, \lambda\right)}_{\vect{\Theta}| \vect{Z} = \vect{z}} \right) & \text{ if } \lambda \in \set{K}_{Q,\vect{z}}\\
+\infty & \text{ otherwise,}
\end{array}
\right. \quad
\end{IEEEeqnarray}
where the function $\mathsf{R}_{\vect{z}}$ is defined in~\eqref{EqRxy} and the measure $P^{\left(Q, \lambda\right)}_{\vect{\Theta}| \vect{Z} = \vect{z}}$ is the solution to the ERM-RER  problem in~\eqref{EqERMRER}.
The sensitivity of the expected empirical risk $\mathsf{R}_{\vect{z}}$ due to a deviation from $P^{\left(Q, \lambda\right)}_{\vect{\Theta}| \vect{Z} = \vect{z}}$ to $P$ is $\mathsf{S}_{Q, \lambda}\left( \vect{z}, P\right)$.
\end{definition}
\subsection{Dataset-Dependent Bounds}
The following theorem introduces an upper bound on the sensitivity.

\begin{theorem}\label{TheoremSensitivityA}
Given a $\sigma$-finite measure $Q \in \triangle\Bormeaspace{\set{M}}$ and a dataset $\vect{z} \in \left( 
\set{X} \times \set{Y}\right)^{n}$, it holds that, for all $\lambda \in \set{K}_{Q, \vect{z}}$, with $\set{K}_{Q,\vect{z}}$ in~\eqref{EqSetKxy}, and for all probability measures $P \in \triangle_{Q}\Bormeaspace{\set{M}}$, 
\begin{IEEEeqnarray}{rcl}
\abs{\mathsf{S}_{Q, \lambda}\left( \vect{z}, P \right)  }& \leqslant &  \sqrt{2 B_{Q,\vect{z}}^2 D\left( P \| P^{\left(Q, \lambda\right)}_{\vect{\Theta}| \vect{Z} = \vect{z}} \right) },
\end{IEEEeqnarray}
where  the function $\mathsf{S}_{Q, \lambda}$ is defined in~\eqref{EqDefSensitivity}; and  the constant $B_{Q, \vect{z}} \in [0,+\infty)$ is
\begin{IEEEeqnarray}{rcl}
\label{EqB}
B_{Q, \vect{z}}^2 = \sup_{\gamma \in \set{K}_{Q, \vect{z}}}  K^{(2)}_{Q, \vect{z}}\left(-\frac{1}{\gamma} \right),
\end{IEEEeqnarray}
with $K^{(2)}_{Q, \vect{z}}$ being the second derivative of the function  $K_{Q,\vect{z}}$ in~\eqref{EqK}.
\end{theorem}
\begin{IEEEproof}
The proof is presented in \cite[Appendix~V]{InriaRR9454}.
\end{IEEEproof}
In Theorem~\ref{TheoremSensitivityA}, the second derivative of the function $K_{Q, \vect{z}}$ in~\eqref{EqK} plays a central role. The function $K_{Q, \vect{z}}$ is continuous and differentiable infinitely many times in $\left( -\infty, 0 \right)$ \cite[Lemma $2.8$]{InriaRR9454}. Moreover, from \cite[Lemma $2.10$]{InriaRR9454}, it follows that if $\vect{\Theta}$ is the random vector  that induces the measure $P^{\left(Q, \lambda\right)}_{\vect{\Theta}| \vect{Z} = \vect{z}}$ in~\eqref{EqGenpdf}, with $\lambda \in \set{K}_{Q,\vect{z}}$, the empirical risk $\mathsf{L}_{\vect{z}}$ in~\eqref{EqLxy} becomes the real random variable  
$\mathsf{L}_{\vect{z}}\left( \vect{\Theta} \right)$
whose mean,  variance, and third cumulant are respectively $K^{(1)}_{\vect{z}}\left(-\frac{1}{\lambda}\right)$, $K^{(2)}_{\vect{z}}\left( -\frac{1}{\lambda} \right)$, and~$K^{(3)}_{\vect{z}}\left( -\frac{1}{\lambda} \right)$. 

Theorem~\ref{TheoremSensitivityA} establishes an upper and a lower bound on the increase and decrease of the expected empirical risk  that can be obtained by deviating from the optimal solution of the ERM-RER in~\eqref{EqERMRER}. More specifically, note that for all  probability measures $P \in \triangle_{Q}\Bormeaspace{\set{M}}$, it holds that,
\begin{IEEEeqnarray}{rcl}
\label{EqCaireNecokly}
 \mathsf{R}_{\vect{z}}\left( P \right)    & \geqslant& \mathsf{R}_{\vect{z}}\left( P^{\left(Q, \lambda\right)}_{\vect{\Theta}| \vect{Z} = \vect{z}} \right) - \sqrt{2 B_{Q, \vect{z}}^2 D\left( P \| P^{\left(Q, \lambda\right)}_{\vect{\Theta}| \vect{Z} = \vect{z}} \right) }, \mbox{ and } \,\quad \\
 \label{EqCaireNecoklyA}
 \mathsf{R}_{\vect{z}}\left( P \right)    & \leqslant&   \mathsf{R}_{\vect{z}}\left( P^{\left(Q, \lambda\right)}_{\vect{\Theta}| \vect{Z} = \vect{z}} \right) + \sqrt{2 B_{Q, \vect{z}}^2 D\left( P \| P^{\left(Q, \lambda\right)}_{\vect{\Theta}| \vect{Z} = \vect{z}} \right) } . \quad
\end{IEEEeqnarray}

The following theorem highlights the fact that the measure that minimizes the expected empirical risk subject to a constraint in the relative entropy with respect to the ERM-RER optimal measure  $P^{\left(Q, \lambda\right)}_{\vect{\Theta}| \vect{Z} = \vect{z}}$ in~\eqref{EqGenpdf} is also the solution to an ERM-RER problem with parameters $Q$ and $\omega$, for some specific $\omega>0$. 
\begin{theorem}\label{TheoremSensitivityB}
Given a $\sigma$-finite measure $Q \in \triangle\Bormeaspace{\set{M}}$,  a dataset $\vect{z} \in \left( 
\set{X} \times \set{Y}\right)^{n}$, and a nonnegative real  $\lambda \in \set{K}_{Q, \vect{z}}$, with $\set{K}_{Q,\vect{z}}$ in~\eqref{EqSetKxy}, consider the following optimization problem  
\begin{subequations}\label{EqImprovementp235a8}
\begin{IEEEeqnarray}{rcl}
\min_{P \in \triangle_{Q}\Bormeaspace{\set{M}}} & &  \int \mathsf{L}_{\vect{z}}(\vect{\theta}) \mathrm{d} P (\vect{\theta}),\\
\text{subject to:} & \quad & D\left( P\|P^{\left(Q, \lambda\right)}_{\vect{\Theta}| \vect{Z} = \vect{z}} \right) \leqslant c,
\end{IEEEeqnarray}
\end{subequations}
with,  
$c$ denoting a nonnegative constant;
$P^{\left(Q, \lambda\right)}_{\vect{\Theta}| \vect{Z}=\vect{z}}$  the probability measure  in~\eqref{EqERMRER}; and  
$\mathsf{L}_{\vect{z}}$ the function in~\eqref{EqLxy}.
Then, the solution to the optimization problem in~\eqref{EqImprovementp235a8} is a probability measure $P^{\left(Q, \omega\right)}_{\vect{\Theta}| \vect{Z} = \vect{z}}$ satisfying for all $\vect{\theta} \in \supp P$,
\begin{IEEEeqnarray}{rcl}\label{EqGenSolB}
\frac{\mathrm{d}P^{\left(Q, \omega\right)}_{\vect{\Theta}| \vect{Z} = \vect{z}}}{\mathrm{d}Q} \left( \vect{\theta} \right) 
  & =& \exp\left( - K_{Q,\vect{z}}\left(- \frac{1}{\omega} \right) - \frac{1}{\omega} \mathsf{L}_{\vect{z}}\left( \vect{\theta}\right)\right),
\end{IEEEeqnarray}
 with $\omega \in \left( 0,  \lambda\right]$ such that
\begin{IEEEeqnarray}{rcl}
\label{EqBombasticElastic}
D\left( P^{\left(Q, \omega\right)}_{\vect{\Theta}| \vect{Z} = \vect{z}}  \| P^{\left(Q, \lambda\right)}_{\vect{\Theta}| \vect{Z} = \vect{z}} \right) = c.
\end{IEEEeqnarray}
\end{theorem}
\begin{IEEEproof}
The proof is presented in \cite[Appendix~W]{InriaRR9454}. 
\end{IEEEproof}
 
 \subsection{Dataset-Independent Bounds}
 
Consider a probability measure, denoted by $P_{\vect{Z}} \in \triangle\bigg( \left( \set{X} \times \set{Y} \right)^{n}, \left( \mathscr{F}\left(\mathcal{X}\times\mathcal{Y}\right) \right)^n \bigg)$, such that for all  $\set{A} \in \big( \mathscr{F}\left(\mathcal{X}\times\mathcal{Y}\right) \big)^n$ of the form $\set{A} = \set{A}_1 \times \set{A}_2 \times \ldots \times \set{A}_n$ with $\set{A_i} \in  \mathscr{F}\left(\mathcal{X}\times\mathcal{Y} \right)$  and $i \in \lbrace 1,2, \ldots, n \rbrace$, it holds that 
\begin{equation}\label{EqProdPXY}
P_{\vect{Z}}\left( \set{A} \right) = \prod_{t=1}^{n} P_{X Y} \left( \set{A}_t \right),
\end{equation}
where the probability measure $P_{X Y}$ is defined in~\eqref{EqPxy}.
More specifically, $P_{\vect{Z}}\left( \set{A} \right)$ is the probability measure induced by a random variable $\vect{Z} = \left( \left( X_{1}, Y_{1} \right), \left( X_{2}, Y_{2} \right), \ldots, \left( X_{n}, Y_{n} \right) \right)$, in which the $n$ random variables $\left( X_{1}, Y_{1} \right), \left( X_{2}, Y_{2} \right), \ldots, \left( X_{n}, Y_{n} \right)$ are independent and identically distributed according to $P_{X Y}$. 

Let the set $\set{K}_{Q}$, with $Q \in \triangle\Bormeaspace{\set{M}}$, be 
\begin{equation}
\label{EqSetKQ}
\set{K}_{Q} =  \bigcap_{\vect{z} \in \supp P_{\vect{Z}} } \set{K}_{Q, \vect{z}},
\end{equation}
where the set $\set{K}_{Q, \vect{z}}$ is defined in~\eqref{EqK} and the probability measure $P_{\vect{Z}}$ is defined in~\eqref{EqProdPXY}.
The set $\set{K}_{Q}$ in~\eqref{EqSetKQ} can be empty for some choices of the $\sigma$-finite measure $Q$ and empirical loss function $\mathsf{L}_{\vect{z}}$ in~\eqref{EqLxy}. Nonetheless, from \cite[Lemma~$2.2$]{InriaRR9454}, it follows that when $Q$ is a probability measure, then,
\begin{equation}
\set{K}_{Q}= \left( 0, +\infty \right).
\end{equation}
Using this notation, the following corollary of Theorem~\ref{TheoremSensitivityA} provides an upper bound on the expectation of the sensitivity with respect to the probability measure $P_{\vect{Z}}$ in~\eqref{EqProdPXY}.

\begin{corollary}\label{LemmaBoomBoomUp}
Given a $\sigma$-finite measure $Q \in \triangle\Bormeaspace{\set{M}}$, for all $\lambda \in \set{K}_{Q}$, with $\set{K}_{Q}$ in~\eqref{EqSetKQ}, and for all probability measures $P \in \triangle_{Q}\Bormeaspace{\set{M}}$, it holds that 
\begin{IEEEeqnarray}{rcl}
\nonumber
\int \hspace{-1ex}\abs{\mathsf{S}_{Q, \lambda}\left( \vect{z}, P \right)  } \mathrm{d} P_{\vect{Z}}(\vect{z}) 
& \leqslant &   \int \hspace{-1ex} \sqrt{2 B_{Q,\vect{z}}^2 D\left( P\| P^{\left(Q, \lambda\right)}_{\vect{\Theta}| \vect{Z} = \vect{z}} \right) }\mathrm{d} P_{\vect{Z}}(\vect{z}) ,\\
\label{EqEhihAou}
\end{IEEEeqnarray}
where $B_{Q,\vect{z}}$ is defined in~\eqref{EqB};  the probability measure $P^{\left(Q, \lambda\right)}_{\vect{\Theta}| \vect{Z} = \vect{z}}$ is the solution to the ERM-RER  problem in~\eqref{EqERMRER}; and the probability measure $P_{\vect{Z}}$ is defined in~\eqref{EqProdPXY}.
\end{corollary}

In the following theorem, the expectation of the sensitivity with respect to the measure $P_{\vect{Z}}$ in~\eqref{EqProdPXY} is shown to have an upper bound that can be expressed in terms of the lautum information between the models and the data sets.
\begin{theorem}\label{LemmaBoomBoomDown}
Given a $\sigma$-finite measure $Q \in \triangle\Bormeaspace{\set{M}}$, for all $\lambda \in \set{K}_{Q}$, with $\set{K}_{Q}$ in~\eqref{EqSetKQ}, it holds that 
\begin{IEEEeqnarray}{l}
\nonumber
\int \hspace{-1ex}\abs{\mathsf{S}_{Q, \lambda}\left( \vect{z}, P^{\left(Q, \lambda\right)}_{\vect{\Theta}} \right)  } \mathrm{d} P_{\vect{Z}}(\vect{z})  \\
\label{EqFreeOfLace}
 \leqslant   \sqrt{2 B_{Q}^2  \int D\left( P^{\left(Q, \lambda\right)}_{\vect{\Theta}}  \| P^{\left(Q, \lambda\right)}_{\vect{\Theta}| \vect{Z} = \vect{u}} \right)\mathrm{d} P_{\vect{Z}}(\vect{u}) },
\end{IEEEeqnarray}
where the probability measure $P^{\left(Q,\lambda\right)}_{\vect{\Theta}| \vect{Z} = \vect{z}}$ is the solution to the ERM-RER  problem in~\eqref{EqERMRER}; the probability measure $P_{\vect{Z}}$ is defined in~\eqref{EqProdPXY}; the probability measure $P^{\left(Q, \lambda \right)}_{\vect{\Theta}}$ is such that for all $\set{A} \in \BorSigma{\set{M}}$,
\begin{equation}
\label{EqBarPThetaX}
P^{\left(Q, \lambda\right)}_{\vect{\Theta}}\left( \set{A} \right) = \int P^{\left(Q, \lambda\right)}_{\vect{\Theta} | \vect{Z} = \vect{z}  } \left( \set{A} \right)  \mathrm{d}P_{\vect{Z}}\left( \vect{z} \right);
\end{equation} 
and the constant $B_{Q}$ satisfies
\begin{equation}
\label{EqBB}
B_{Q}^2 = \sup_{\vect{z} \in \supp P_{Z}}  B^2_{Q,\vect{z}},
\end{equation}
with $B_{Q, \vect{z}}$ defined in~\eqref{EqB}.
\end{theorem}
\begin{IEEEproof}
The proof is presented in \cite[Theorem $3.5$]{InriaRR9454}. 
\end{IEEEproof}
Given a $\sigma$-finite measure $Q \in \triangle\Bormeaspace{\set{M}}$ and a positive real $\lambda \in \set{K}_{Q}$, with $\set{K}_{Q}$ in~\eqref{EqSetKQ},
let $\vect{Z}$ and $\vect{\Theta}$ be the random variables that jointly induce a probability measure $P^{\left( Q, \lambda \right)}_{\vect{Z} \vect{\Theta}}$ with marginals $P_{\vect{Z}}$ in \eqref{EqProdPXY} and $P^{\left(Q, \lambda\right)}_{\vect{\Theta}}$ in \eqref{EqBarPThetaX}.
Under these assumptions, the right-hand side in~\eqref{EqFreeOfLace} can be written in terms of the lautum information \cite{palomar2008lautum} between the random variables $\vect{Z}$ and $\vect{\Theta}$, which is denoted by $\mathtt{L}\left( \vect{Z}; \vect{\Theta}  \right)$. More specifically, note that 
\begin{equation}
\mathtt{L}\left( \vect{Z}; \vect{\Theta}  \right) = \int  D\left( P^{\left(Q, \lambda\right)}_{\vect{\Theta}}  \| P^{\left(Q, \lambda\right)}_{\vect{\Theta}| \vect{Z} = \vect{z}} \right)\mathrm{d} P_{\vect{Z}}(\vect{z}).
\end{equation}
In a nutshell, it can be concluded that the expectation of $\abs{\mathsf{S}_{Q, \lambda}\left( \vect{z}, P^{\left(Q, \lambda\right)}_{\vect{\Theta}} \right)}$ with respect to the measure $P_{\vect{Z}}$ in~\eqref{EqProdPXY} is upper bounded by the lautum information between the random variables $\vect{Z}$ and $\vect{\Theta}$, which represent the datasets and the models, respectively. 
\balance 
\section{Final Remarks}\label{SecDiscussions}
 
This work focuses on a special case of the ERM problem in which the random variables $X$ and $Y$ in~\eqref{EqTheModel} are such that $Y$ is deterministic given the ground truth model $\vect{\theta}^{\star}$ and a realization of the random variable $X$. That is, all data points in the dataset $\vect{z}$ in~\eqref{EqTheDataSet} are pairs of patterns correctly labeled.  
A more practical case is that in which the random variables $X$ and $Y$ satisfy $Y = f\left( \vect{\theta}^{\star}, X\right) + W$, for some random variable $W$, which represents an additive noise.  In this case, data points in the dataset $\vect{z}$ in~\eqref{EqTheDataSet} are pairs of patterns and labels that are not necessarily correct (in the sense of~\eqref{EqTheModel}).  Nonetheless, the results presented in this paper can be extended to such a  case. 
\bibliographystyle{IEEEtran}
\bibliography{references}
\balance
\end{document}